\documentclass{article}
\usepackage{amssymb}
\usepackage{amsmath}
\usepackage{amsfonts}
\usepackage{graphicx}
\usepackage{subcaption}
\usepackage{tabularx}
\usepackage{booktabs}
\usepackage{multirow}
\usepackage{appendix}

\usepackage[preprint]{corl_2025} 

\title{Improving Trajectory Stitching with Flow Models}

\author{
  Reece O'Mahoney\\
  Oxford Robotics Institute\\
  University of Oxford\\ 
  \texttt{reeceo@robots.ox.ac.uk} \\
  \And
  Wanming Yu\\
  Oxford Robotics Institute\\
  University of Oxford\\
  \texttt{wanming@robots.ox.ac.uk} \\
  \And
  Ioannis Havoutis\\
  Oxford Robotics Institute\\
  University of Oxford\\
  \texttt{ioannis@robots.ox.ac.uk} \\
}

\begin{document}
\maketitle

\begin{abstract}
     Generative models have shown great promise as trajectory planners, given their affinity to modeling complex distributions and guidable inference process. Previous works have successfully applied these in the context of robotic manipulation but perform poorly when the required solution does not exist as a complete trajectory within the training set. We identify that this is a result of being unable to plan via \textit{stitching}, and subsequently address the architectural and dataset choices needed to remedy this. On top of this, we propose a novel addition to the training and inference procedures to both stabilize and enhance these capabilities. We demonstrate the efficacy of our approach by generating plans with out of distribution boundary conditions and performing obstacle avoidance on the Franka Panda in simulation and on real hardware. In both of these tasks our method performs significantly better than the baselines and is able to avoid obstacles up to four times as large. Code and videos are available at \url{https://reeceomahoney.vercel.app/publications/flow_planner}
\end{abstract}



\section{Introduction}

Generative modeling has shown significant advances in recent years, especially in the domains of image and video generation~\citep{rombach2022highresolution, podell2023sdxlimprovinglatentdiffusion, esser2023structurecontentguidedvideosynthesis}. This success has also led to considerable research in its application to control tasks \cite{janner2022planning, ajay2023conditional, chi2024diffusion}. Flow matching is a recent generative modeling paradigm that is similar to diffusion ~\citep{ho2020denoising, song2020generative}, but is conceptually much simpler and in recent works has demonstrated state of the art performance in multiple domains~\cite{esser_scaling_2024, polyak2025moviegencastmedia, le2023voiceboxtextguidedmultilingualuniversal, black__0_2024}. Both of these techniques fall under the unified family of methods known as Generator Matching (GM)~\cite{holderrieth2025generatormatchinggenerativemodeling}. In this work we opt for the latter framework due to its faster convergence, quicker inference, and better asymptotic performance.

Previous works have used these approaches for robot manipulation~\cite{chi2024diffusion, reuss2023goal, wang2023diffusion} but primarily for short horizon action generation. Two other works explore guided long horizon planning~\cite{carvalho_motion_2024, luo2024potentialbaseddiffusionmotion}, but we note a major drawback when deploying these method ourselves. While guided diffusion is extremely capable at steering generation to select the best complete trajectory from the dataset, it is largely unable to create new trajectories from novel compositions of sub-trajectories. Without this second mode, for a given initial and final state pair, generation is limited to existing trajectories in the training dataset that already connect these points. As a result, a high density of initial-final state pairs, each exhibiting all the variations that would be desired at test time, is required to produce flexible behavior. 
For a model capable of stitching, the combinatorial space of clips needed for this same level of capability reduces exponentially. Exploring how to plan via \textit{stitching}, as opposed to merely \textit{selecting} is the focus of this work. In particular, we identify three key deficiencies that have prevented previous models from developing robust stitching capabilities. 

\begin{figure}
    \centering
    \includegraphics[width=\linewidth]{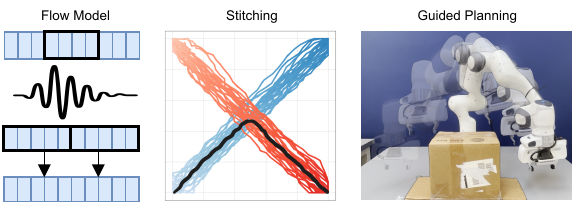}
    \caption{Overview: We use a flow model with three key components: a local receptive, field, dataset augmentation with out addition, and trajectory splitting at train time and test time. This allows us to have significant improved stitching capabilities, and more flexible guided planning, leading an ability to avoid much larger obstacles than previous methods.}
    \label{fig:overview}
\end{figure}

Firstly, we find that model architecture is extremely important to ensure each timestep has a local ``receptive field". This term has been previously used to describe the range of states that each neuron is activated by in a CNN~\cite{luo2017understandingeffectivereceptivefield}, but here we use it to describe the range of states that each \textit{state can attend to}. We find that if the conditioning or inference process is not completely local with respect to time, each state will be biased by information from past and future states into merely continuing what has been seen in the dataset for this same set of states and as a result, cannot handle novel conditioning combinations. We explore which architectural choices allow this capability in section \ref{sec:local}.

Secondly, we identify that proper dataset augmentation is required for GM models to stitch well. In particular, we find that action noise addition is uniquely effective when compared to other common domain randomization schemes. We hypothesize this is due to the breaking of the correlation between the agent's states, something unique to the noise addition scheme. We explore this further in section \ref{sec:aug}. 

Lastly, we find that while the above two additions help stitching, they are not sufficient for a full and reliable solution. The models they produce still suffer from two problems. Firstly, while the models initially fit the conditioning under increasingly strong guidance, over the course of training they tend to ``overfit" the unguided solution produce inconsistent plans. Secondly, for stronger guidance and large deviations, the amount of inconsistency in the trajectory grows, especially at the endpoints. Surprisingly, we find that a similar technique can solve both of these problems, depending on whether it is applied during training or inference. We call this ``trajectory splitting" which we explain in section \ref{sec:split}.

We use the combination of all of these techniques to propose a model we call Flow Planner (FP).
We demonstrate the ability of our method to stitch different trajectories by training it on simple cross-shaped dataset of motions on the Franka Panda and measuring its success in generating feasible plans between novel start and end points. We additionally demonstrate the flexibility of our approach for guided generation in an obstacle avoidance task in sim and on hardware, and show the ability to avoid much larger obstacles than previous methods.


\section{Related work}
\label{sec:related-work}

\subsection{Diffusion models for planning}

Using diffusion models for planning was first introduced in \cite{janner2022planning} but has since been explored in numerous works \cite{ubukata2024diffusionmodelplanningsystematic, pearce2023imitatinghumanbehaviourdiffusion, he2023diffusionmodeleffectiveplanner} and contexts such as robotic manipulation \cite{chi2024diffusion, reuss2023goal, carvalho_motion_2024, wang2024pocopolicycompositionheterogeneous, luo2024potentialbaseddiffusionmotion}, locomotion \cite{huang2024diffuseloco, omahoney2025offlineadaptationquadrupedlocomotion}, and offline RL \cite{wang2023diffusion}. Additional capabilities have been explored on top of regular diffusion including classifier-free guidance \cite{ajay2023conditional}, and refining \cite{lee2023refiningdiffusionplannerreliable}. \cite{black__0_2024} was the first work to apply flow matching to robotic manipulation, noting its high precision and multi-modality. The closest work to ours is Motion Planning Diffusion \cite{carvalho_motion_2024}, but with the primary difference being they are unable to stitch trajectories, and so most of their obstacle avoidance capabilities come from having access to the obstacles during dataset collection, with minor deviations on top of this. In contrast we are able to perform large deviations from our unguided plans with no privileged information. Additionally, while we both use the same architecture, we provide a mechanistic justification for this choice that previous works have not explored.

\subsection{Trajectory Stitching}

Trajectory stitching is a popular subject of interest, with several previous works exploring it in the context of diffusion \cite{li2024diffstitchboostingofflinereinforcement, kim2024stitchingsubtrajectoriesconditionaldiffusion, lee2024gtagenerativetrajectoryaugmentation, luo2025generativetrajectorystitchingdiffusion}, model-based planning \cite{char2022batsbestactiontrajectory, hepburn2022modelbasedtrajectorystitchingimproved}, and transformers \cite{wu2023elasticdecisiontransformer}. Stitching is also mentioned briefly in \cite{janner2022planning} but not expanded on in any meaningful sense. While these works explore a similar problem to us, they do not propose any hypothesis similar to local receptive fields and tie this to architecture and conditioning choices. While the authors of \cite{luo2025generativetrajectorystitchingdiffusion} do explore similar ideas to our splitting strategy, this is in the context of composing shorter clips into longer ones as opposed to aiding guidance.


\section{Background}
\label{sec:background}

\subsection{Flow matching}

Flow Matching (FM)~\cite{lipman2024flowmatchingguidecode} builds a probability path $(p_t)_{0 \leq t \leq 1}$, from a known source distribution $p_0 = p$ to the data target distribution $p_1 = q$. To do this, an ODE is defined via a time-dependent vector field $u:[0,1] \times \mathbb{R}_d \rightarrow \mathbb{R}_d$ which in turn determines a time-dependent flow $\psi:[0,1] \times \mathbb{R}_d \rightarrow \mathbb{R}_d$, defined as:
\begin{equation}
\label{eq:flow}
   \frac{d}{dt} \psi_t(x) = u_t(\psi_t(x)), 
\end{equation}
where $\psi_t := \psi(t, x)$ and $\psi_0(x) = x$. The velocity field $u_t$ generates the probability path $p_t$ if its flow $\psi_t$ satisfies:
\begin{equation*}
    X_t := \psi_t(X_0) \sim p_t \text{ for } X_0 \sim p_0.
\end{equation*}
One popular choice of probability path is the conditional optimal-transport path since it minimizes the kinetic energy of the velocity field. This is defined as:
\begin{equation*}
X_t = tX_1 + (1 - t)X_0 \sim p_t,
\end{equation*}
where ${p_0 = \mathcal{N}(x \mid 0, I)}$.
To sample from this path we can then train a neural network parameterized velocity field $u_t^\theta$ to regress to a target velocity field $u_t$ known to generate the desired probability path $p_t$, but this is usually intractable as it requires marginalizing over the entire training set. The objective simplifies drastically by conditioning the loss on a single target sample, and remarkably, as shown in~\citep{lipman2023flowmatchinggenerativemodeling}, these full and conditional flow matching losses provide the same gradients to learn $u_t^\theta$. By solving the conditional form of Eq \ref{eq:flow} we can then formulate the full conditional Flow Matching objective:
\begin{equation*}
\mathcal{L}^{\text{OT,Gauss}}_{\text{CFM}} (\theta) = \mathbb{E}_{t,X_0,X_1} \| u^\theta_t (X_t) - (X_1 - X_0) \|^2, \text{ where  } t \sim \mathcal{U}[0, 1].
\end{equation*}
To generate samples from the target distribution $X_1$ we approximate the solution to the flow ODE by starting from some initial sample from $X_0$ and integrating using the Euler method:
\begin{equation*}
    X_{t+h} = X_t + hu_t(X_t).
\end{equation*}

\subsection{Guidance}

The literature in diffusion models shows that guided model generation by direct condition is most effective in applications where a large amount of target samples $X_1$ share the same guiding signal $Y$ , such as in class guidance~\cite{nichol2022glidephotorealisticimagegeneration}. However, guiding is more challenging in settings where the guidance variable Y is non-repeating and complex, such as image captions. For flows trained with Gaussian paths, classifier guidance~\cite{song_score-based_2021, dhariwal_diffusion_2021} can be applied by utilizing the transformation between velocity fields and score functions for conditional distributions shown in \cite{zheng2023guidedflowsgenerativemodeling}:
\begin{equation*}
    u^{\theta,\phi}_t(x \mid y) 
    = u^\theta _t (x) + b_t \nabla \log p^\phi_{Y \mid t}(y \mid x),
\end{equation*}
where $b_t=\frac{1-t}{t}$ for the optimal transport path. Although not theoretically justified, we found better results in practice using $b_t=1-t$.


\section{Flow Planner}

In this section we describe the three techniques that enable Flow Planner's robust stitching capabilities. First we expand on our hypothesis of local receptive fields and explain the implications of this on architecture choice. Then we discuss action noise addition and theorize the reason for its superior performance over other randomization schemes. Lastly we look at the mode collapse and dynamics inconsistency that occur in guided planning and explain how we fix these both with trajectory splitting.

\subsection{Local receptive fields}
\label{sec:local}

\begin{figure}[h]
    \centering
    \begin{subfigure}[b]{0.3\textwidth}
        \centering
        \includegraphics[width=\linewidth]{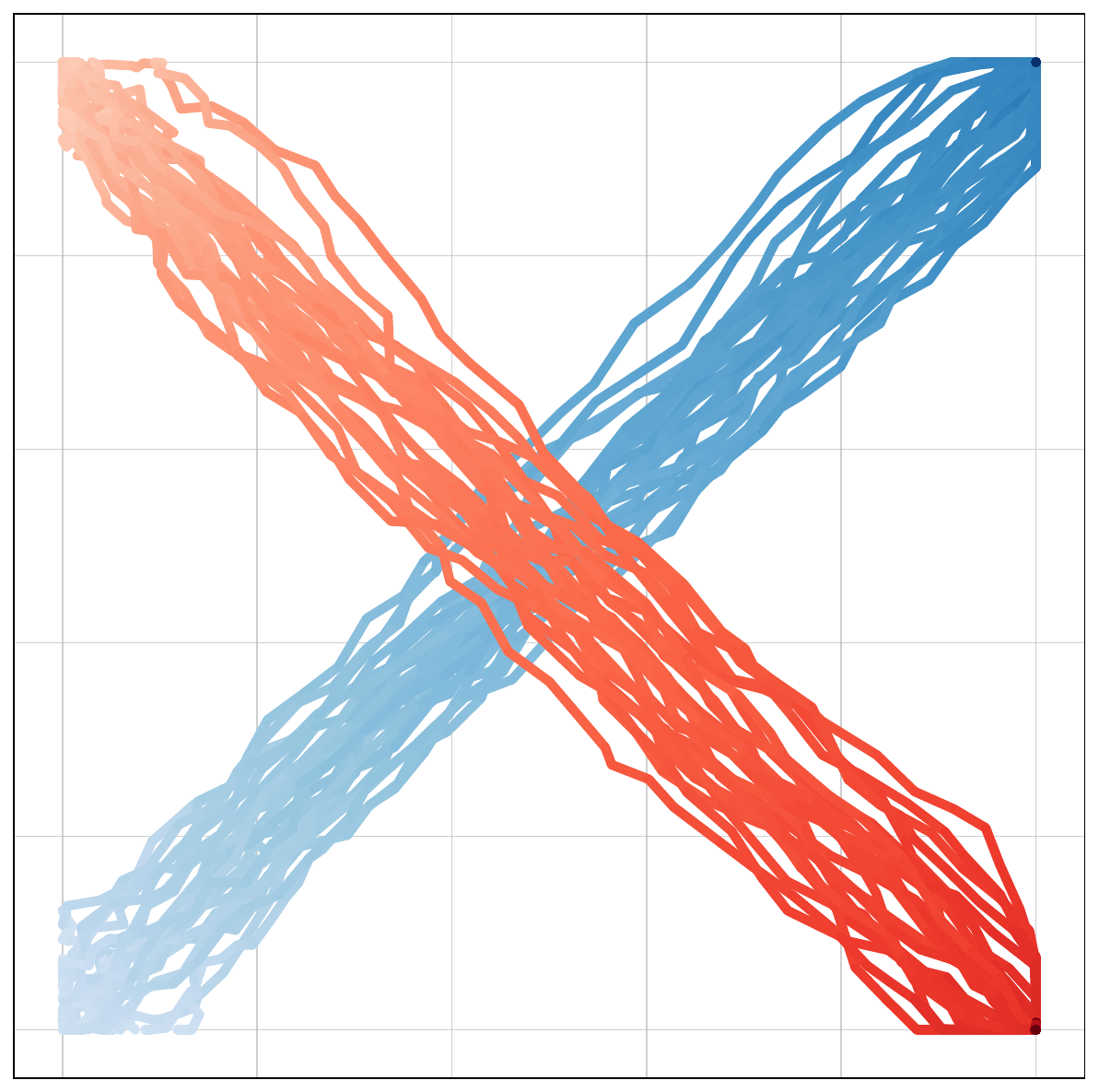}
        \caption{Original dataset}
        \label{fig:particle_dataset}
    \end{subfigure}
    \hfill
    \begin{subfigure}[b]{0.3\textwidth}
        \centering
        \includegraphics[width=\linewidth]{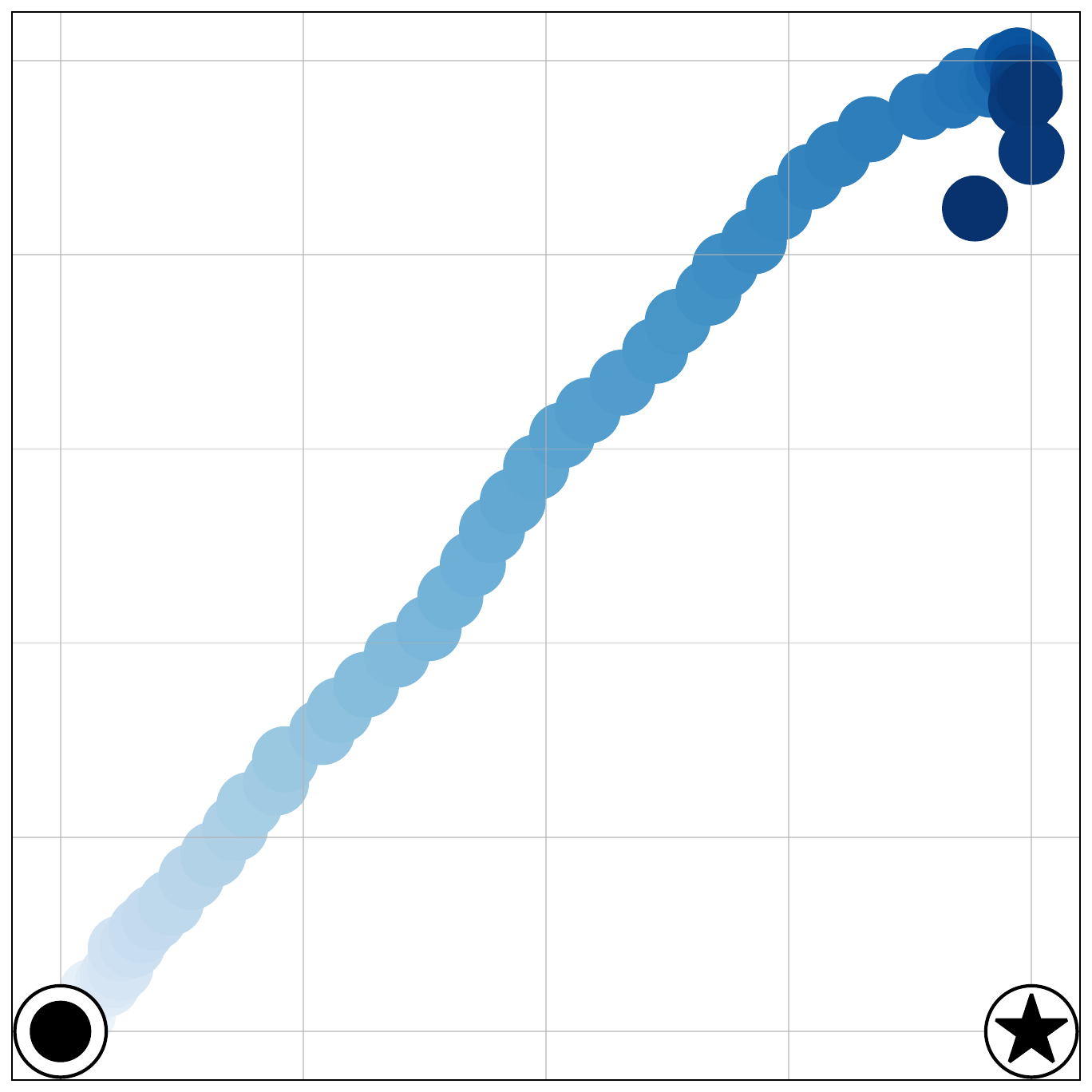}
        \caption{Stitching failure}
        \label{fig:particle_no_stitch}
    \end{subfigure}
    \hfill
    \begin{subfigure}[b]{0.3\textwidth}
        \centering
        \includegraphics[width=\linewidth]{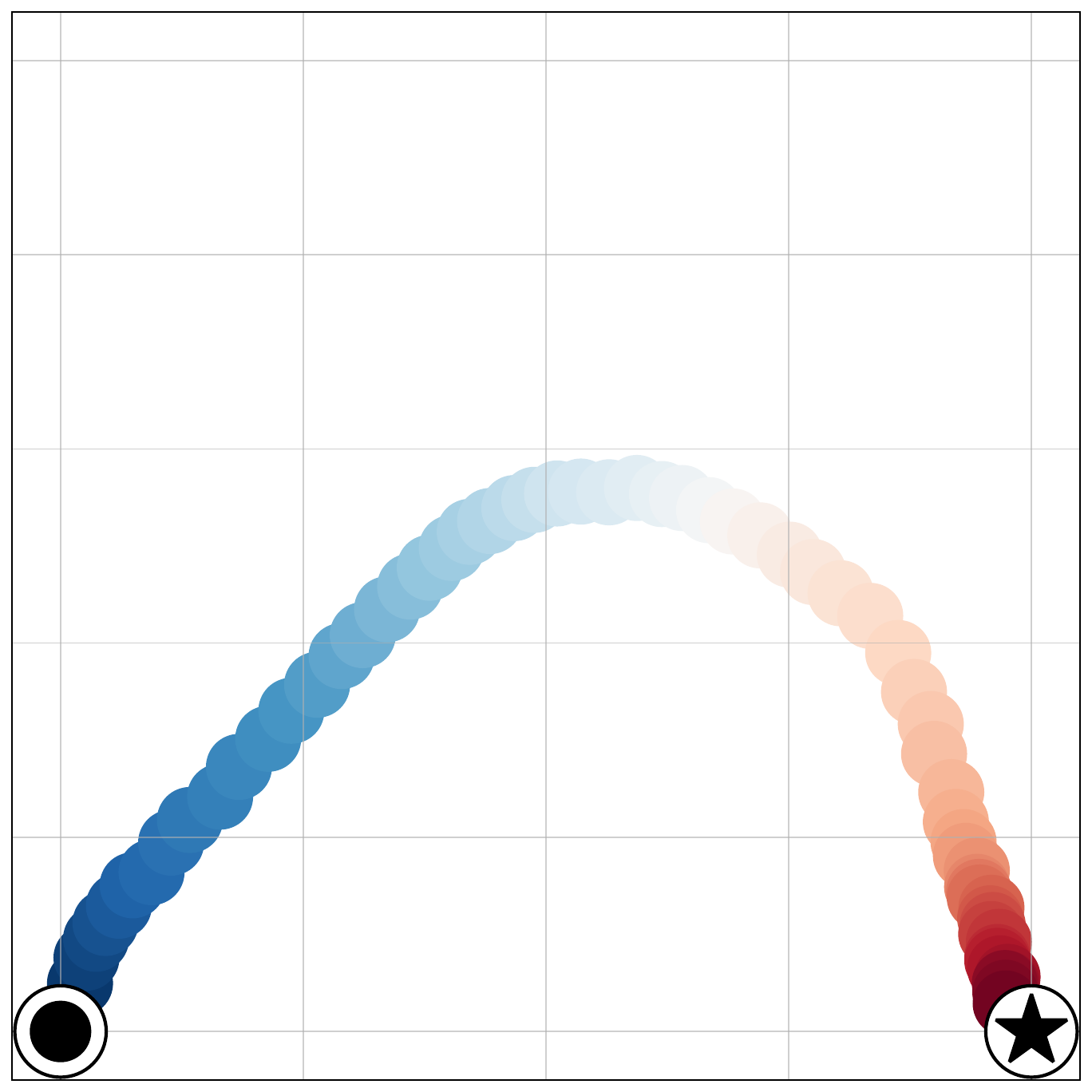}
        \caption{Stitching success}
        \label{fig:particle_stitch}
    \end{subfigure}
    \caption{Demonstration of local receptive fields: (a) Original dataset (b) Transformer with direct conditioning i.e. non-local information is unable to reconcile novel conditioning (c) UNet with inpainting conditioning that is completely local and hence able to stitch. The circle and star show the initial and final conditioning states.}
    \label{fig:attn_leakage}
\end{figure}

A crucial aspect to enabling stitching capabilities is ensuring that models have a \textit{local receptive field}, i.e. that each state only has access to the the information from the other states in its local neighborhood, as opposed to the plan's overall initial or goal positions. The intuition behind this is as follows.

The loss function for any generative model, either directly or indirectly, aims to maximize the conditional likelihood of the samples it generates given the training distribution. When this conditioning information is non-local at training time, this incentives learning a representation that is strongly biased towards reproducing whole samples from the training set. Even if only a subset of this information is given as test time, e.g. just the start and end states, we observe that this bias still persists. We verify this with an environment used in \cite{janner2022planning} consisting of a 2D particle that moves in a cross-shape as shown in Fig \ref{fig:particle_dataset}. If this information is globally accessible during training, the model just learns to reproduce whole pre-existing samples, and as such using a novel conditioning pair at test time like in Fig \ref{fig:particle_no_stitch} yields inconsistent results.

In contrast to this, a model with a local receptive field, should only be aiming to maximize the likelihood of each state given its local neighborhood, which for trajectory planning translates to maximizing local consistency, with no preferences regarding the global structure this produces. As such, the resulting model should be able to handle novel conditioning information much more flexibly, and stitch together segments of trajectories that together satisfy the overall boundary conditions. Again this agrees with our experimental findings, with Fig \ref{fig:particle_stitch} showing a trajectory produced from a model with only local conditioning.

In terms of concrete architectural choices, we identify two models -- transformers and UNets --, and two conditioning schemes -- direct and inpainting -- that are common in the GM literature~\cite{chi2024diffusion, janner2022planning}. For each of these, the former results in a global receptive field, whereas the latter, a local one. We provide a quantitative evaluation of the stitching performance of each on the Franka Panda in section \ref{sec:stitch_results}.

\subsection{Dataset augmentation with action noise}
\label{sec:aug}

\begin{figure}[h]
    \centering
    \begin{subfigure}[b]{0.45\textwidth}
        \centering
        \includegraphics[width=\linewidth]{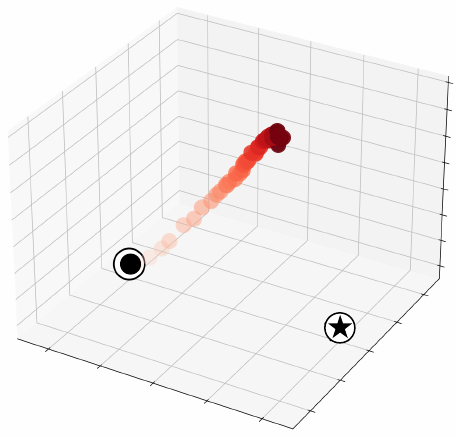}
        \caption{No noise addition}
        \label{fig:franka_no_noise_x}
    \end{subfigure}
    \begin{subfigure}[b]{0.45\textwidth}
        \centering
        \includegraphics[width=\linewidth]{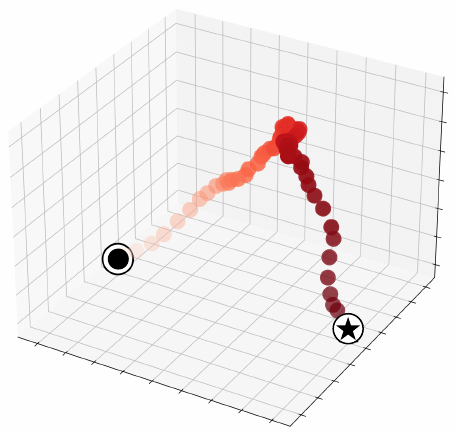}
        \caption{Noise addition}
        \label{fig:franka_noise_x}
    \end{subfigure}
    \caption{Effect of dataset augmentation: The two figures above show the end-effector trajectories of a flow model controlling a Franka Panda. The dataset forms a cross shape similarly to Fig. \ref{fig:attn_leakage}. (a) No action noise during dataset collection and leads to simply selecting an existing trajectory. (b) Action noise addition during dataset generation enables the stitching of the different modes.}
    \label{fig:dataset_aug}
\end{figure}

We observe that the second crucial element to our model's stitching performance is dataset augmentation. In particular, what we observed to be uniquely effective was adding Gaussian noise to our collection policy's actions during dataset generation. Fig \ref{fig:dataset_aug} shows the end-effector trajectories generated by a model trained on a dataset of Franka motions that form a cross shape similar to the previous section. The left model's dataset had no augmentation, whereas the right model's did. Again similarly to the previous section, when given a novel conditioning pair, only the latter is able to stitch a viable trajectory from the two modes it has seen. Domain randomization techniques are well known in the RL literature to improve generalization~\cite{tobin2017domainrandomizationtransferringdeep}, but what we found to be surprising was the unique effectiveness of this method of randomization over others. In section \ref{sec:stitch_results} we perform a quantitative comparison against two other methods, randomizing initial and goal positions, and random force injection, both of which are ineffective. We hypothesize that this is due the breaking of the correlations between the joint states that only action noise produces. The reasoning for this is as follows. 

In order for a model to be able to stitch trajectory segments, it needs to have a high density of overlapping trajectories that it can ``jump" between. Just having these physically overlap in joint space however is not a sufficient criteria. We propose that if the correlation between the individual joint states within these overlapping segments is still too strong, the model will overfit to these particular patterns, thus preventing it from jumping to other nearby clips. While randomizing initial and goal states and force perturbation induces this first criteria, they do not induce the second i.e., given some local history of states, the proceeding states will still largely be the same. It is the state-wise independence of action noise addition that enables this second criteria and thus allows for robust stitching. To test this further we also test applying the same randomly sampled noise value to every joint in section \ref{sec:stitch_results}. This is also ineffective, further supporting our hypothesis.

\subsection{Trajectory splitting}
\label{sec:split}

\begin{figure}[h]
    \centering
    \begin{subfigure}[b]{0.3\textwidth}
        \centering
        \includegraphics[width=\linewidth]{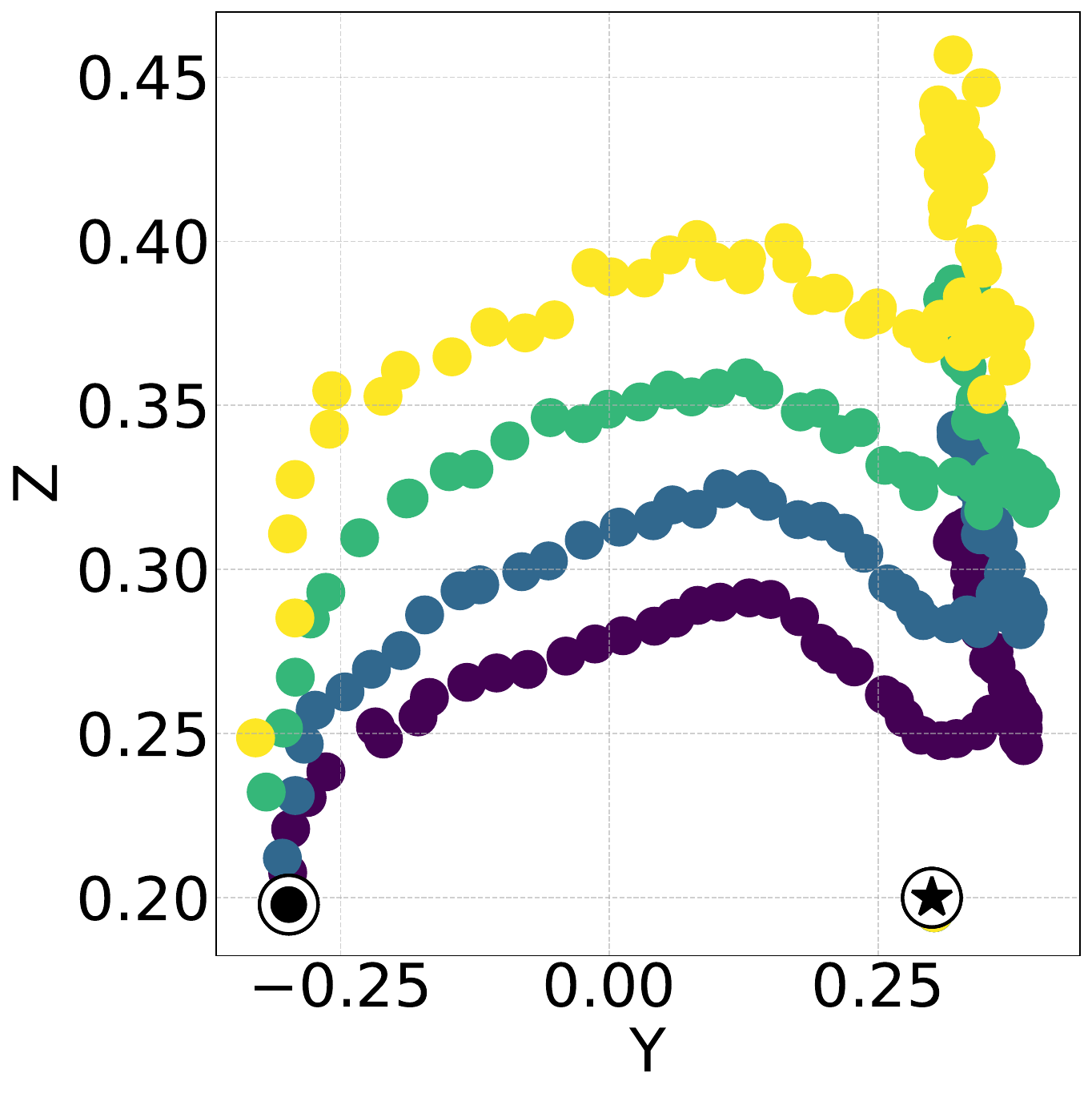}
        \caption{Mode collapse}
        \label{fig:mode_collapse}
    \end{subfigure}
    \hfill
    \begin{subfigure}[b]{0.3\textwidth}
        \centering
        \includegraphics[width=\linewidth]{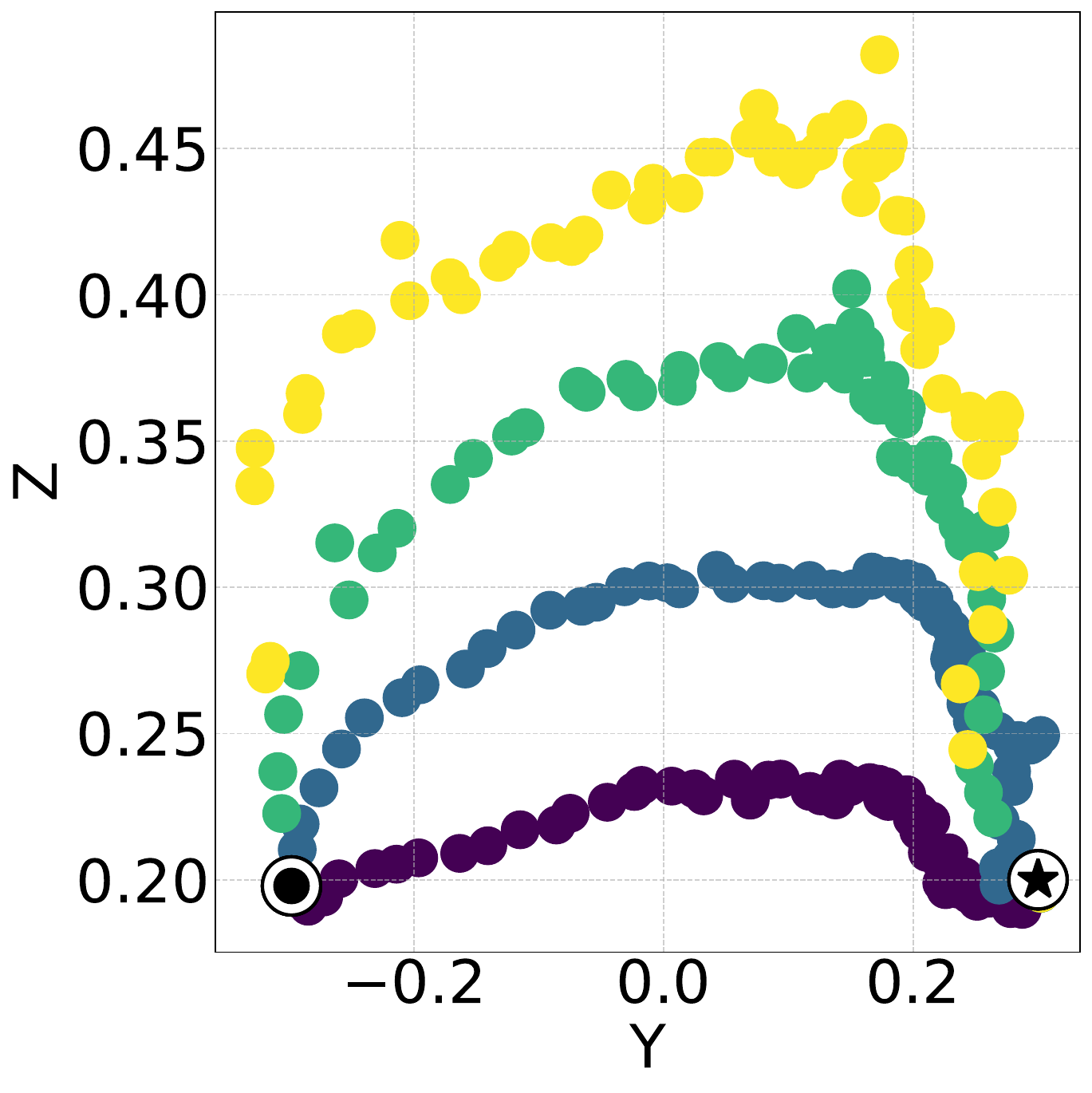}
        \caption{Without splitting}
        \label{fig:no_split}
    \end{subfigure}
    \hfill
    \begin{subfigure}[b]{0.3\textwidth}
        \centering
        \includegraphics[width=\linewidth]{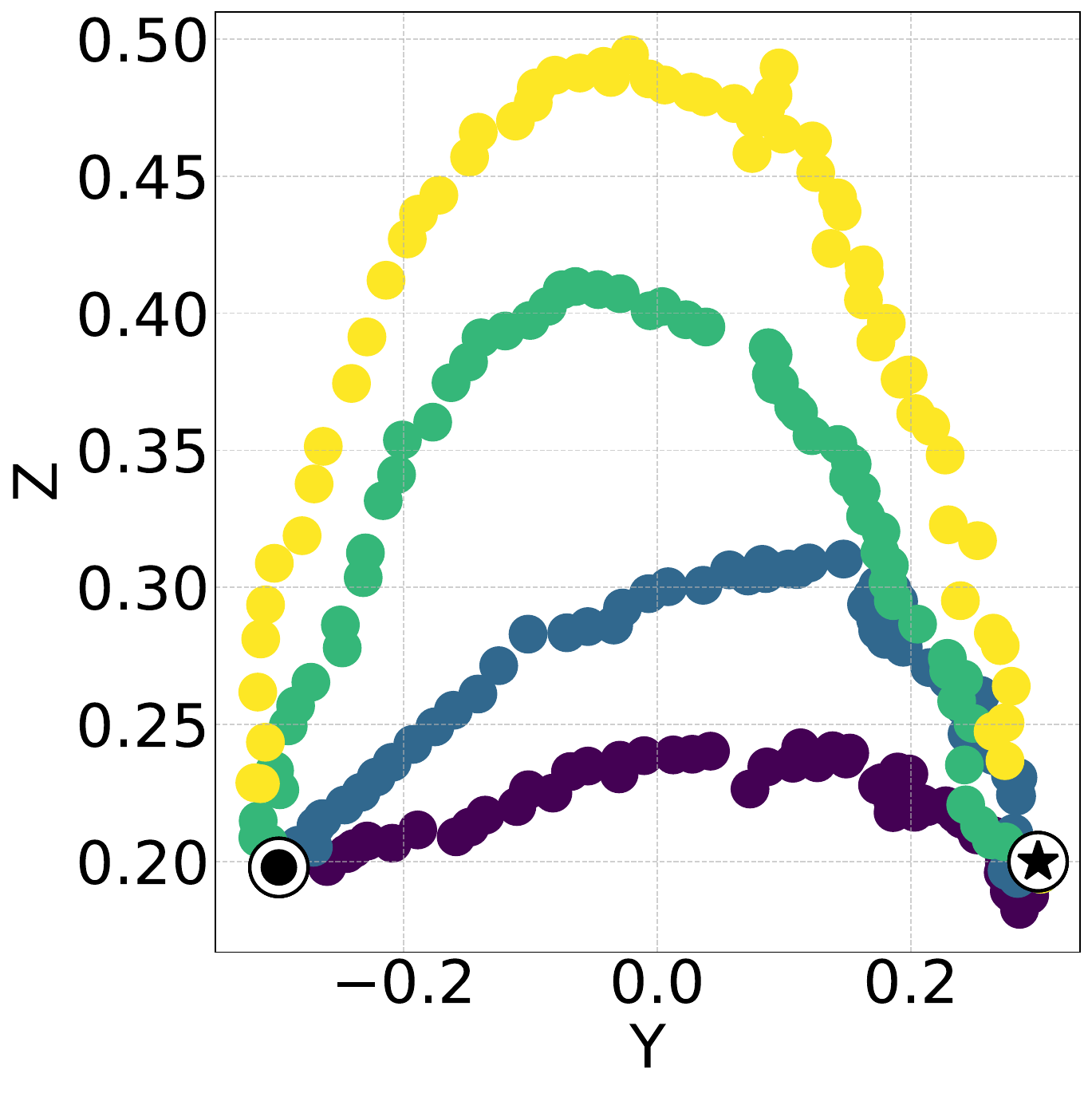}
        \caption{With splitting}
        \label{fig:split}
    \end{subfigure}
    \caption{Trajectory splitting: (a) Training without train time trajectory splitting causes mode collapse. (b) Train time splitting prevents mode collapse. (c) Inference time splitting increases maximum consistent deviation.}
    \label{fig:splitting}
\end{figure}

While the addition of the techniques from the previous two sections massively increase stitching performance, we find that the resulting models still suffer from two major drawbacks. To demonstrate this we setup a demonstration model trained on a dataset of Franka Panda trajectories that follow mostly straight lines in task space. For a guidance function, we use the same analytical scheme described in \cite{carvalho_motion_2024} consisting of signed distance function between each link and the obstacle, calculated via differential forward kinematics, and a Gaussian process smoothness cost. The obstacle here is just a single point between the start and end positions. The plots of the end-effector trajectories produced by this model are shown in Fig \ref{fig:splitting}.

Firstly, while the initial models during training produce consistent plans at various guidance scales, we observe that this isn't a stable phenomenon, and over the course of training, the models experience a kind of mode collapse as shown in Fig \ref{fig:mode_collapse}, where the produced trajectory no longer bends to try satisfy the conditioning, but instead just translates. Secondly, we observe that as the guidance scale increases, so does the dynamics inconsistency of the resulting trajectories. This is an inevitable consequence of the trade-off between consistency and cost minimization that guided planning can be interpreted as \cite{janner2022planning}, but the rate at which this occurs limits the potential flexibility of our planner. Surprisingly, we observe that a similar mechanism can both prevent this first weakness, and dramatically slow this second, depending on whether it is applied during training or inference. We call this technique \textit{trajectory splitting} and it consists of the following. 

At train time, instead of just training with full length trajectories, we simply mix half-length trajectories into our batches with a 50\% probability. We find that it is important that the trajectory lengths remain powers of 2 due to the particular structure of the UNet --- arbitrary lengths can result in order of magnitude increases in error. This simple technique completely eliminates the previously explained mode collapse. Using a similar logic to our previous sections, we hypothesize that this is due to variation in trajectory length reducing the overfitting effect of the model, giving it more flexibility to deviate from the unguided solution. 

At inference time, trajectory splitting is implemented by adding an additional step after the model's usual denoising loop. We take the ``initial guess" produced by this first process, re-noise it up to 50\% with newly sampled noise, and split it in halves. We then perform a second denoising loop with 50\% as many steps on each of these halves separately, using the midpoint as the missing boundary condition for each half. Figures \ref{fig:no_split} and \ref{fig:split} show the the resulting trajectories without and with this process, visually demonstrating the reduction in dynamics inconsistency experienced at the same level of guidance. The result of these modifications is greater obstacle avoidance capabilities, which we demonstrate empirically in section \ref{sec:guided_planning}.

\section{Experimental Results}
\label{sec:results}

In the following section we empirically demonstrate the benefits of our approach on two different measures: stitching ability, and obstacle avoidance. We test both of these on the Franka Panda platform in simulation and on real hardware. We collect our dataset by rolling out an inverse dynamics controller across 10,000 parallel simulation environments for episodes of 64 timesteps.

\subsection{Stitching}
\label{sec:stitch_results}

To measure the stitching ability of each approach, we design a benchmark as follows. We generate a dataset forming an cross shape, like the one used in section \ref{sec:aug}. At training time the model only sees opposite side conditioning pairs. At test time, we generate a trajectory by randomly sampling one of the same side conditioning pairs. We then record the mean squared error between the first and last \textit{planned} states as the \textit{stitching error} to quantify the stitching ability -- since the true first and last are inpainted -- and the true initial and final states. We average these results over batches of 64, each with 64 steps and report them below in tables \ref{tab:stitching-arch} and \ref{tab:stitching-data}. The details of the different methods are as follows.

\textbf{FP (Ours)} is our method. \textbf{FP w/o inpainting} is our method with direct conditioning. \textbf{Transformer} is our method with a transformer substituted for the UNet. \textbf{Transformer w/o inpainting} is the same as the previous but with direct conditioning. \textbf{MPD} is the baseline approach from \cite{carvalho_motion_2024}. \textbf{Same noise} applied the same Gaussian noise value to all the joints. \textbf{Random pos} randomized the start and goal positions in some small range around the target. \textbf{Random forces} adds randomly sampled force perturbations to the end-effector. All Gaussian noise is applied directly to the torques sent to the joints. Our method clearly performs best in both tests with minimum average stitching error. The transformer with inpainting being the next best architecture choice, and random positions the next best for dataset augmentation.

\renewcommand{\arraystretch}{1.25}
\begin{table*}[h]
    \centering
    \begin{tabularx}{\linewidth}{*{5}{>{\centering\arraybackslash}X}}
        \toprule
        FP (Ours) & FP w/o inpainting & Transformer & Transformer w/o inpainting & MPD \\
        \midrule
        $ \mathbf{0.16} \pm 0.07 $ & $ 0.71 \pm 0.24 $ & $ 0.30 \pm 0.09 $ & $ 1.73 \pm 0.49 $ & $1.00 \pm 0.97$ \\
        \bottomrule
    \end{tabularx}
    \caption{Effects of architecture choice on stitching error}
    \label{tab:stitching-arch}
\end{table*}

\begin{table*}[h]
    \centering
    \begin{tabularx}{\linewidth}{*{6}{>{\centering\arraybackslash}X}}
        \toprule
        FP & Same noise & No noise & Random pos & Random forces & MPD \\
        \midrule
         $ \mathbf{0.16} \pm 0.07 $ & $ 1.62 \pm 0.25 $ & $ 0.81 \pm 0.05 $ & $ 0.54 \pm 0.06 $ & $ 0.85 \pm 0.23 $ & $1.00 \pm 0.97$ \\
        \bottomrule
    \end{tabularx}
    \caption{Effects of dataset augmentation on stitching error}
    \label{tab:stitching-data}
\end{table*}
\renewcommand{\arraystretch}{1}

\subsection{Guided Planning}
\label{sec:guided_planning}

\begin{figure}[h]
    \centering
    \begin{subfigure}[b]{0.4\textwidth}
        \centering
        \includegraphics[width=\linewidth]{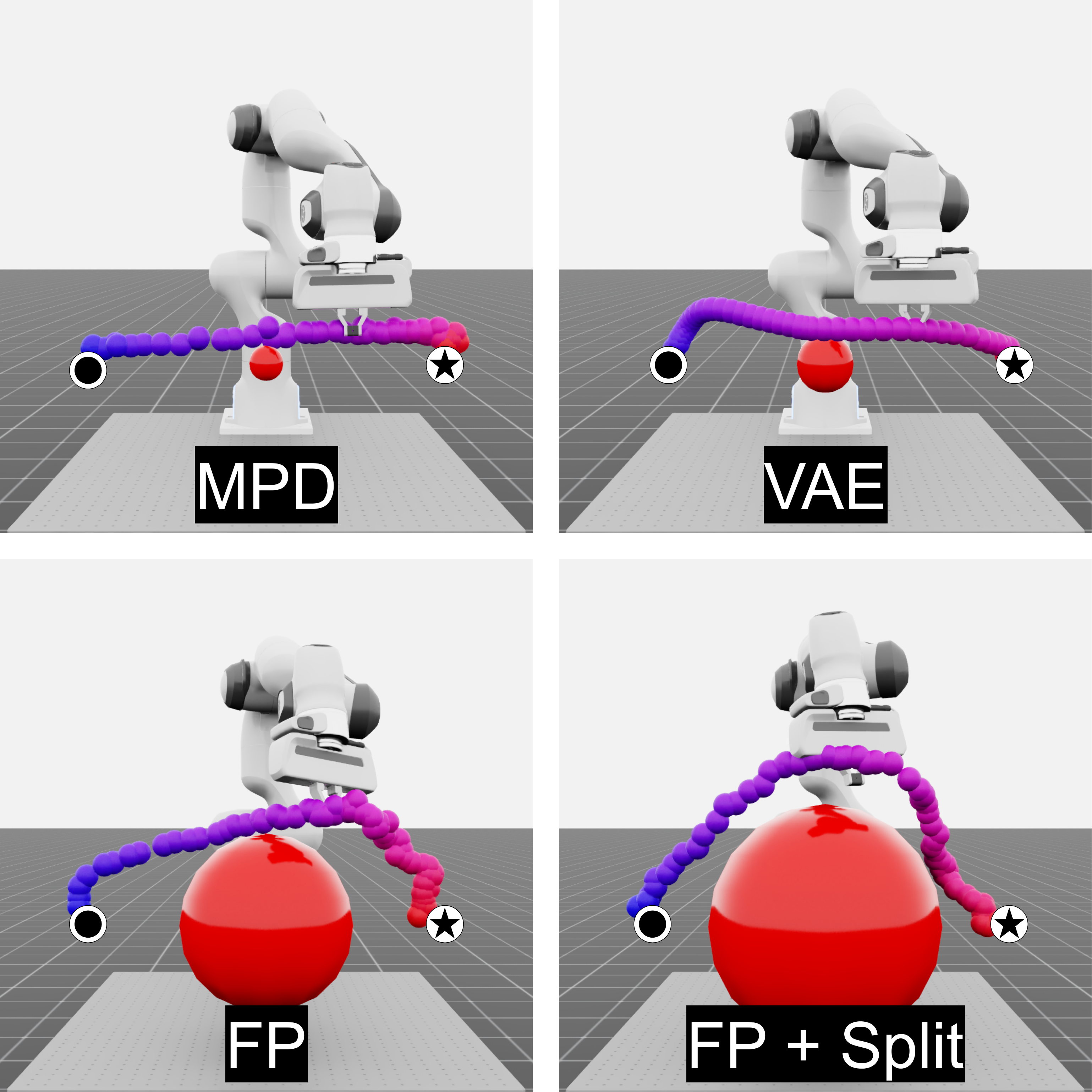}
    \end{subfigure}
    \begin{subfigure}[b]{0.55\textwidth}
        \centering
        \includegraphics[width=\linewidth]{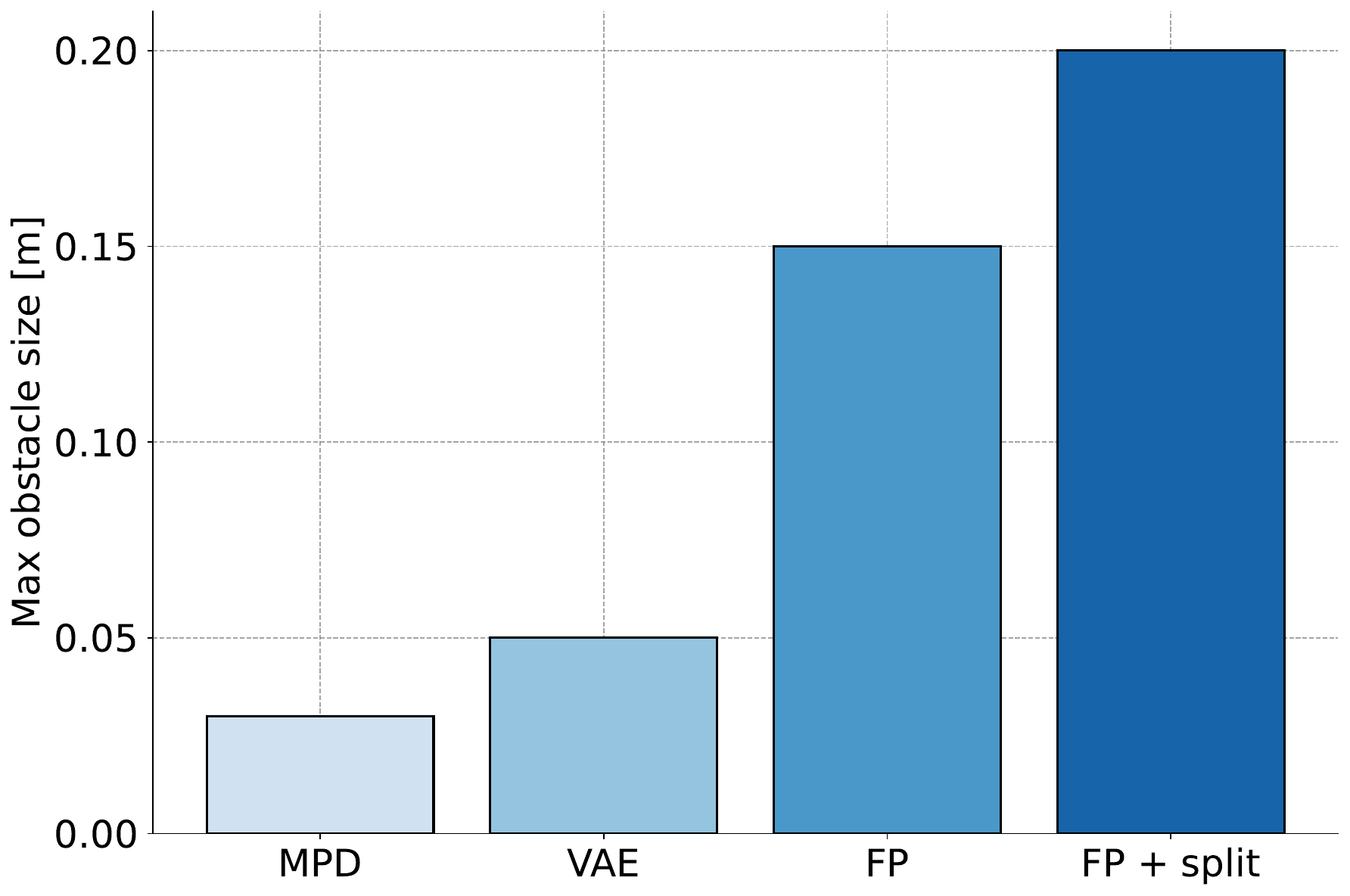}
    \end{subfigure}
    \caption{Performance benchmark in an obstacle avoidance task on a Franka arm in simulation. Left: Sample trajectories from obstacle avoidance experiments. Right: Maximum obstacle radius that each method was able to reliably avoid.}
    \label{fig:guided_planning}
\end{figure}

To quantitatively compare the flexibility of our planner to other approaches, we design an obstacle avoidance task where the Franka arm must navigate over some spherical obstacle placed between its initial and goal positions. We use a sphere for the same reasons as the authors of MPD, being that it's the simplest signed distance function to compute. This method would be compatible with arbitrary shapes and multiple obstacles. Since the dataset trajectories are almost completely straight, there is no possible pre-existing single example in the training set able to solve this task, hence, stitching is required. Our guidance function is the same used by the authors of \cite{carvalho_motion_2024} as described in our method. We found the collision and smoothness terms to be sufficient and had no benefit from adding the additional ones. Sample trajectories and numerical results of these experiments are shown in Fig \ref{fig:guided_planning}. The numbers in the right figure are the maximum object radius that the arm was able to reliably avoid. The \textbf{MPD baseline} is unable to stitch at all and hence performed the worst, barely being able to deviate around a $3$ cm sphere. The \textbf{VAE (Variational autoencoders) baseline} uses activation maximization \cite{Hung_2022} with gradients from our analytical guidance function. This performed slightly better but it is worth noting that it is a local planner, and hence is more susceptible to myopic failure modes than the others. Lastly, \textbf{FP} and \textbf{FP + split} are our method without and with inference time trajectory splitting respectively. This showed the best obstacle avoidance results by a large margin, with splitting enhancing performance even further.

\begin{figure}[h]
    \centering
    \includegraphics[width=\linewidth]{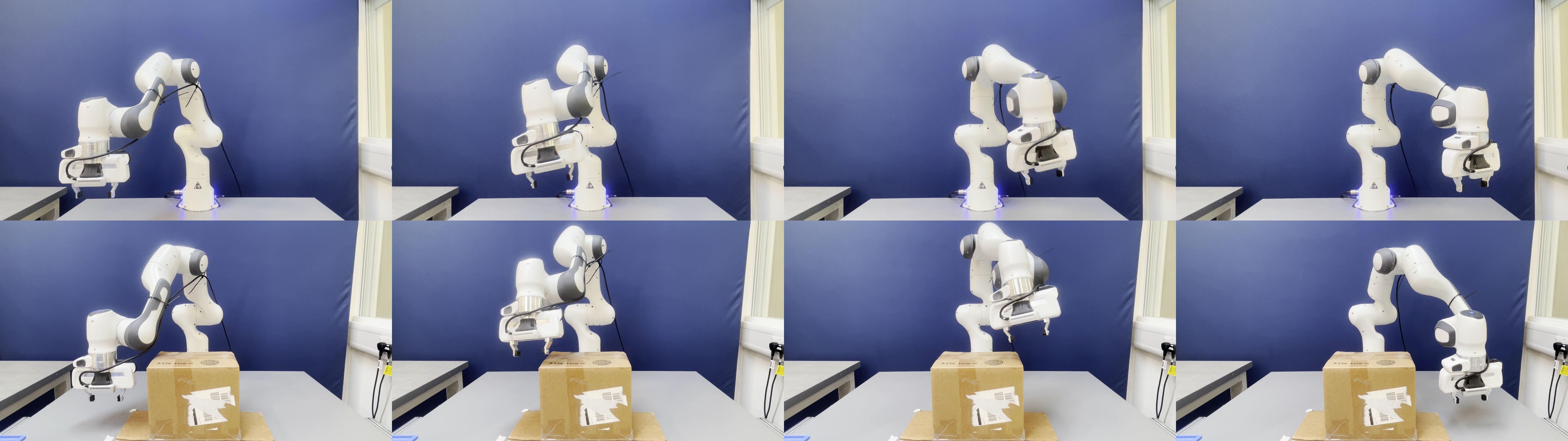}
    \caption{Hardware deployment, the top row shows an unguided plan and the bottom a guided one.}
    \label{fig:hardware}
\end{figure}

Lastly to verify the validity, we deploy the same obstacle avoidance task as above on a hardware, we use a box that can be contained within the maximum size sphere to demonstrate the generalizability of our proposed method. This is shown in Fig \ref{fig:hardware}. The top rows shows a plan without guidance that simply translates across in a straight line, whereas the bottom shows one with guidance, which avoids the obstacle while still reaching the commanded final position. To control the arm, we send the planned joint positions and velocities to an onboard PD controller, which smooths them out with a moving average.

\section{Conclusion}
\label{sec:conclusion}

In conclusion, we present an approach to guided planning for robotic locomotion with a massively improvement flexibility to previous methods by focusing on developing robust stitching capabilities. We do by analyzing the architectural choices that bias models towards planning by selecting, comparing various data augmentation schemes and suggesting a novel addition to both training and inference that stabilize and enhance these capabilities. Our method outperforms our baselines on both generating out of distribution trajectory plans and guided obstacle avoidance. We lastly test the robustness of these results by deploying on a real Franka Panda manipulator.


\section{Limitations}

A significant limitation of our work was recovering the original behavior from the noised dataset during deployment. We found that there was a direct trade-off between smooth trajectories at low noise levels, and robust stitching at high noise levels. We tried many other less destructive augmentation schemes but none seemed to provide the robustness of noise addition. Because of this, during hardware deployment, a combination of a Savitzky–Golay filter and an moving average had to be employed to re-smooth the trajectories. A interesting direction for future work would be how to enable the same level of stitching in a less destructive manor.

Another direction attempted during this work was the application of novel masking schemes to the transformer models in order to induce a local receptive field. We found that transformers exhibited faster convergence and better accuracy than UNets and so would be a desirable alternative. One theoretically sound approach seemed to be a local attention mask, that only used the main and 2 off diagonals, but we found this to be ineffective in practice. Solving this issue would be another interesting direction for future work.


\clearpage


\bibliography{main}  


\clearpage
\appendix

\section{Hyperparameters}

\renewcommand{\arraystretch}{1.5}
\begin{table}[h]
    \centering
    \begin{tabular}{|c|c|}
        \hline
        Parameter & Value  \\
        \hline
        UNet dims & [32, 64, 128, 256] \\
        Timestep embed dim & 32 \\
        Train steps & 200k \\
        Learning rate & 2e-4 \\
        Inference steps & 10 \\
        Collision cost & 0.1 \\
        Smoothness cost & 1e-6 \\
        \hline
    \end{tabular}
    \medskip
    \caption{Hyperparameters}
    \label{tab:hyperparameters}
\end{table}
\renewcommand{\arraystretch}{1}

\section{Success rate at different obstacle sizes}

\begin{figure}[h]
    \centering
    \includegraphics[width=0.8\linewidth]{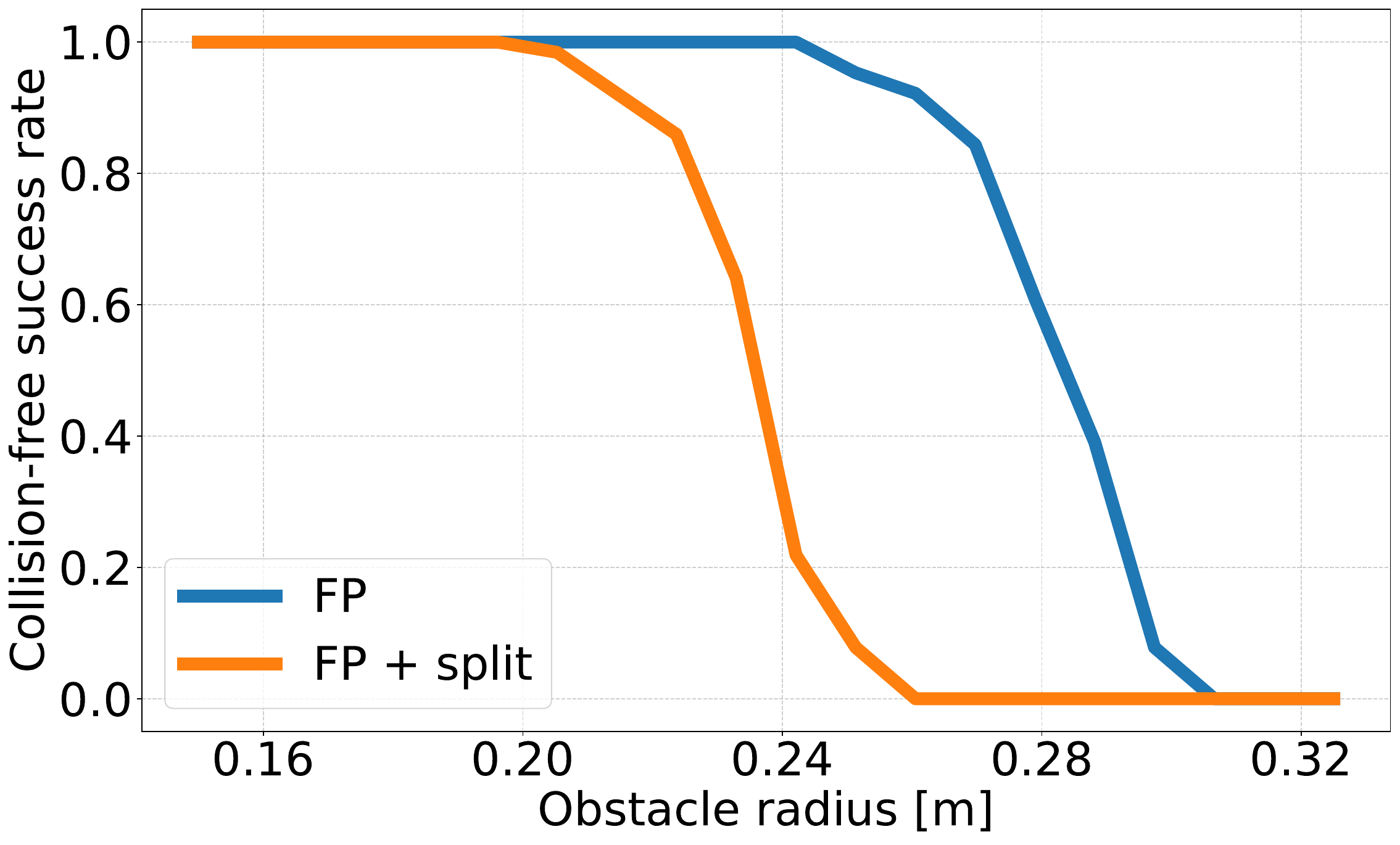}
    \caption{Collision-free success rates at different obstacle sizes}
    \label{fig:success_rates}
\end{figure}

To provide a more detailed definition of the term ``reliably avoid" used in section \ref{sec:guided_planning} we recorded the success rates of the generated plans with different sized obstacles in the same setup shown previously. These results are shown in Fig \ref{fig:success_rates}. Here success is just defined at reaching the goal state without any collisions. The improvement from FP to FP + split is in large part due to being able to use a larger guidance scale without the resulting actions becoming dynamically inconsistent.

\end{document}